\documentclass{ieeetj}
\usepackage{cite}
\usepackage{amsmath,amssymb,amsfonts}
\usepackage[ruled,vlined]{algorithm2e}
\usepackage{graphicx,color}
\usepackage{tikz}
\usetikzlibrary{positioning,arrows.meta,shadows}
\usepackage{textcomp}
\usepackage{xcolor}
\usepackage{multirow}
\usepackage{rotating}
\usepackage{siunitx}
\definecolor{inputcolor}{RGB}{150,190,255}    % stronger blue
\definecolor{hiddencolor}{RGB}{190,150,255}   % stronger purple
\definecolor{embedcolor}{RGB}{150,255,150}   % stronger green
\definecolor{outputcolor}{RGB}{255,180,150}  % stronger orange/pink
\definecolor{arrowcolor}{RGB}{0,0,0}         % keep black
\definecolor{shadowcolor}{gray}{0.4}         % slightly darker shadow

\usepackage{fix-cm}
\usepackage{hyperref}
\hypersetup{
    colorlinks=true,
    linkcolor=NavyBlue,
    citecolor=tmlcncolor,
    urlcolor=NavyBlue
}

\usepackage{booktabs}      % For \toprule, \midrule, \bottomrule
\usepackage[table]{xcolor} % For \rowcolor
\usepackage{pifont}        % For \ding symbols

\usepackage{array}
\definecolor{bestrow}{rgb}{0.90,0.95,1.0}
\newcolumntype{C}[1]{>{\centering\arraybackslash}p{#1}}
\newcommand{\SafeIncludeGraphics}[2][]{%
  \IfFileExists{#2}{\includegraphics[#1]{#2}}{%
    \fbox{\parbox[c][0.28\textheight][c]{0.88\textwidth}{\centering Missing figure file: \texttt{\detokenize{#2}}}}%
  }%
}

\def\BibTeX{{\rm B\kern-.05em{\sc i\kern-.025em b}\kern-.08em
    T\kern-.1667em\lower.7ex\hbox{E}\kern-.125emX}}
\AtBeginDocument{\definecolor{tmlcncolor}{cmyk}{0.93,0.59,0.15,0.02}\definecolor{NavyBlue}{RGB}{0,86,125}}

\def\OJlogo{}
\def\seclogo{}

\def\authorrefmark#1{\ensuremath{^{\textbf{#1}}}}

\makeatletter
\def\@maketitle{\newpage
  \bgroup%
    \vspace*{-10.5pt}%
    \ifx\@doiinfo\@empty\else\vskip3pt{\doi@font Digital Object Identifier 10.1109/\@doiinfo\par}\fi%
    \vskip14.5pt{\titlefont{\textcolor{tmlcncolor}{\@title}}\par\vskip8.5pt}%
    {\authorfont\@author\par}\vskip4pt{\afffont\@affil}\par\@IEEEspecialpapernotice\vskip5pt%
    \ifx\@corresp\@empty\else{\corfont\@corresp\par}\vskip5.5pt\fi%
    \ifx\@authornote\@empty\else{\afffont\@authornote\par}\vskip27.5pt\fi%
    \centerline{\rule{36pc}{.5pt}}\par\vskip11.5pt%
    \ifvoid\abstractbox\else{\reset@font\box\abstractbox}\fi\par\vskip11.5pt%
    \ifvoid\keybox\else{\reset@font\box\keybox}\fi\par\vskip-32pt%
    \centerline{\rule{36pc}{.5pt}}%
    \par\addvspace{33.5pt}\egroup%
    \let\maketitle\relax
}
\makeatother

\begin{document}

\markboth{}{ALI {\textit{ET AL.}}: \MakeUppercase{Fake Job Post Detection using Centroid-Guided Contrastive Loss}}

\title{Beyond Accuracy: Centroid-Guided Contrastive Loss for Structured Fraudulent Job Posting Detection} \author{Syed Ali Ahmed\authorrefmark{1}, Malaika Raza\authorrefmark{2}, Muhammad Shoaib Siddiqui\authorrefmark{3} (Senior Member, IEEE),\\ and Muhammad Rafi\authorrefmark{4} (Member, IEEE)}
\affil{National University of Computer and Emerging Sciences, Karachi, 75020, Pakistan}
\affil{National University of Computer and Emerging Sciences, Karachi, 75020, Pakistan}
\affil{Faculty of Computer and Information Systems, Islamic University of Madinah, Madinah 42351, Saudi Arabia}
\affil{Department of AI \& DS, National University of Computer and Emerging Sciences, Karachi, 75020, Pakistan} \corresp{\MakeUppercase{Corresponding authors}: \MakeUppercase{Syed Ali Ahmed} (email: k224058@nu.edu.pk) and \MakeUppercase{Muhammad Shoaib Siddiqui} (email: shoaib@iu.edu.sa).} \authornote{Muhammad Shoaib Siddiqui's ORCID is 0000-0002-5656-0416. The authors extend their appreciation to the Deanship of Scientific Research, Islamic University of Madinah, Saudi Arabia, for funding this research work.
}

\begin{abstract}
Fraudulent job posting detection aims to identify job advertisements that are corrupted either through fake content, misleading information, or negative intent, disrupting the online eco-system of job-seekers and employers. Existing studies in this domain lack effective methods to simultaneously achieve high accuracy and meaningful structure of latent-space representations that capture subtleties among fake posts. To this end, we propose Centroid-Guided Contrastive Loss (CGCL), a loss function which unifies classification with densely formulated clustering to consistently reshape latent-space through a centroid-driven top-\(k\) push-and-pull mechanism. The complementary nature of CGCL enables the model to enforce accurate decision boundaries and maintain high clustering compactness, effectively capturing both class separability and latent structure. Extensive experiments demonstrate the state-of-the-art (SOTA) performance of our method on EMSCAD, a public benchmark dataset. The code associated with this work is available at:
\url{https://github.com/ali-ahmed925/CGCL_code/tree/main}

\end{abstract}


\IEEEspecialpapernotice{This work has been submitted to the IEEE for possible publication. Copyright may be transferred without notice, after which this version may no longer be accessible.}

\maketitle
\thispagestyle{empty}
\pagestyle{empty}

\section{INTRODUCTION}

\begin{table*}[!t]
\centering
\caption{Comparison of fake job postings detection methods showing feature extraction techniques, imbalance handling approaches, and performance metrics.}
\label{tab:fake_news_comparison}
\begin{tabular}{l c l c l c}
\toprule
\textbf{Ref} & \textbf{Year} & \textbf{Feature Encoding} & \textbf{Imbalance Handling} & \textbf{Best Performing Model} & \textbf{Accuracy} \\
\midrule
\rowcolor[HTML]{E8F5E9}
\cite{dutta2020fake} & 2020 & Categorical encoding & \textcolor{red}{\ding{55}} & Random Forest Classifier & 98.27 \\
\rowcolor[HTML]{D4E6F1}
\cite{anita2021fake} & 2021 & - & \textcolor{red}{\ding{55}} & BiLSTM & 98.0 \\
\rowcolor[HTML]{D4E6F1}
\cite{keerthana2021accurate} & 2021 & TF-IDF & \textcolor{red}{\ding{55}} & MLP Classifier & 71.0 \\
\rowcolor[HTML]{D4E6F1}
\cite{shibly2021performance} & 2021 & - & \textcolor{red}{\ding{55}} & Decision Forests & 95.4 \\
\rowcolor[HTML]{E8F5E9}
\cite{amaar2022detection} & 2022 & TF-IDF & \textcolor{green}{\ding{51}} & Extra Tree Classifier & 99.9 \\
\rowcolor[HTML]{E8F5E9}
\cite{qayyum2023frd} & 2022 & BiLSTM & \textcolor{red}{\ding{55}} & BiLSTM & 97.21 \\
\rowcolor[HTML]{D4E6F1}
\cite{singh2023fake} & 2023 & - & \textcolor{red}{\ding{55}} & DNN & 98.0 \\
\rowcolor[HTML]{D4E6F1}
\cite{pillai2023detecting} & 2023 & One-hot encoding & \textcolor{green}{\ding{51}} & BiLSTM & 98.71 \\
\rowcolor[HTML]{D4E6F1}
\cite{rathudi2023fake} & 2023 & Word2Vec & \textcolor{green}{\ding{51}} & LSTM & 97.18 \\
\rowcolor[HTML]{E8F5E9}
\cite{anbarasu2024fake} & 2024 & TF-IDF & \textcolor{green}{\ding{51}} & SGD Classifier & 98.6 \\
\rowcolor[HTML]{E8F5E9}
\cite{afzal2024identifying} & 2024 & TF-IDF & \textcolor{green}{\ding{51}} & Extra Tree Classifier & 99.76 \\
\bottomrule
\end{tabular}
\end{table*}

\IEEEPARstart{I}{n} the current age of Intelligence, where internet has deeply transformed our modern day life and social media platforms are straightforwardly accessible, companies nowadays, increasingly rely on electronic means to advertise their job postings and recruitment windows. This digitized approach has not only simplified the application process for job seekers but also accelerated recruitment operations for employers. However, malicious attempts to corrupt the recruitment ecosystem have emerged due to the prevalence of fake job postings surfacing on popular job-hunting platforms. These fraudulent postings are a direct attack on applicant's personal information, exposing them to a range of cyber threats, such as identity theft, financial scams, and privacy breaches \cite{rathudi2023fake}. According to a report published by the \textit{Better Business Bureau}, employment scams ranked as the second most risky scam type in 2023, with a 5.2\% increase in reported incidents and an average reported loss of \$1,995 per victim, up from \$1,500 in 2022 \cite{bbb2022report}. Therefore, detecting such fraudulent postings is of utmost importance to ensure compliance and integrity, and also to safeguard job-seekers from financial and identity-related harms.

Machine learning in recent years has emerged as a highly promising technique across a wide range of domains, from healthcare \cite{habehh2021machine, shailaja2018machine, wiens2018machine} and finance \cite{dixon2020machine, rundo2019machine, ahmed2022artificial, kelly2023financial} to video surveillance \cite{8578776, chalapathy2019deep, pang2021deep} and cybersecurity \cite{sarker2020cybersecurity, xin2018machine, shaukat2020survey, apruzzese2023role}. A very impactful application of machine learning is in the field of Natural Language Processing (NLP), which is well-suited for tasks involving unstructured textual data. This makes it a viable approach for problems like fake job post detection where data is available in natural language format. Several machine learning models have been employed to detect fraudulent samples ranging from traditional models such as Naive Bayes Classifier (NBC), K-Neighbors Classifier (KNNs), Decision Tree Classifier (DTC) and more to advanced deep learning architectures like Long Short Term Memory (LSTMs), and Gated Recurrent Units (GRUs). These models capture temporal context and patterns from sequences of textual input, enabling them to classify anomalous samples as "fraudulent postings".

While choosing the right model is undoubtedly a critical aspect of solving the problem, it is only one part of the broader learning scheme. An often unnoticed and equally essential component of any learning framework is the loss or cost function. A loss function directly influences the model to learn subtle patterns in the data by computing the difference between original target labels and predicted outputs, thereby optimizing the model's parameters during training. Most existing studies rely on standard loss functions like Cross-entropy Loss or Hinge Loss, which while effective in many cases, may not be able to fully capture nuanced class boundaries and inter-class separation in such complex tasks where normal and fraudulent samples' features share subtle similarities.

To this end, we propose a novel \textit{Centroid-Guided Contrastive Loss (CGCL)}, which meaningfully reshapes the high-dimensional latent embedding space by incorporating a centroid-based push-and-pull mechanism that enhances intra-class compactness and inter-class separability. We equip CGCL with a top \textit{k} strategy to select only the \textit{k} farthest in-class samples during the pull phase and the closest cross-class samples during the push phase, ensuring a focused and effective feature refinement. Additionally, CGCL leverages a class-balanced Cross-Entropy Loss that guides the classifier towards more robust decision boundaries, maximizing the discrimination between unique classes for improved generalization and classification performance. We train a Multilayer Perceptron with six hidden layers, each followed by a ReLU activation function--- except for the final hidden layer which omits the activation. We conducted extensive experiments on a publicly available dataset from Kaggle \cite{shivamb_fake_job_posting} to validate the effectiveness of our approach. Our contributions can be summarized as follows:

\begin{enumerate}
    \item We propose a novel \textit{Centroid-Guided Contrastive Loss (CGCL)} that ensures intra-class compactness and inter-class separability through top \textit{k} push-pull mechanism.

    \item We integrate CGCL with class-balanced Cross-Entropy Loss, resulting in more robust and discriminative decision boundaries.

    \item We evaluate our complete framework on a benchmark dataset, demonstrating improved performance in class-imbalance scenarios.
\end{enumerate}

\begin{table*}[!t]
\centering
\caption{Job Dataset Schema: Column Specifications and Data Types}
\label{tab:job_dataset_schema}
\begin{tabular}{l c c l}
\toprule
\textbf{Column Name} & \textbf{Data Type} & \textbf{Non-Null Count} & \textbf{Description} \\
\midrule
\rowcolor[HTML]{E8F5E9}
job\_id & int64 & 17880 & Unique identifier for job \\
\rowcolor[HTML]{D4E6F1}
title & object & 17880 & Job position or role name \\
\rowcolor[HTML]{D4E6F1}
location & object & 17534 & City, state, or country posted \\
\rowcolor[HTML]{D4E6F1}
department & object & 6333 & Hiring department within company \\
\rowcolor[HTML]{D4E6F1}
salary\_range & object & 2868 & Offered compensation range \\
\rowcolor[HTML]{D4E6F1}
company\_profile & object & 14572 & Company background or description \\
\rowcolor[HTML]{D4E6F1}
description & object & 17871 & Job duties and responsibilities \\
\rowcolor[HTML]{D4E6F1}
requirements & object & 15146 & Skills or qualifications needed \\
\rowcolor[HTML]{D4E6F1}
benefits & object & 10637 & Perks or advantages offered \\
\rowcolor[HTML]{E8F5E9}
telecommuting & int64 & 17880 & Remote work allowed (binary) \\
\rowcolor[HTML]{E8F5E9}
has\_company\_logo & int64 & 17880 & Company logo presence (binary) \\
\rowcolor[HTML]{E8F5E9}
has\_questions & int64 & 17880 & Screening questions included (binary) \\
\rowcolor[HTML]{D4E6F1}
employment\_type & object & 14409 & Full-time, part-time, contract, etc. \\
\rowcolor[HTML]{D4E6F1}
required\_experience & object & 10830 & Level of experience required \\
\rowcolor[HTML]{D4E6F1}
required\_education & object & 9775 & Minimum education qualification needed \\
\rowcolor[HTML]{D4E6F1}
industry & object & 12977 & Sector of employment \\
\rowcolor[HTML]{D4E6F1}
function & object & 11425 & Primary role or specialization \\
\rowcolor[HTML]{E8F5E9}
fraudulent & int64 & 17880 & Legitimate or fake posting label \\
\bottomrule
\end{tabular}
\end{table*}

\begin{figure*}[!t]
    \centering
    
    \begin{minipage}{0.4\textwidth}
        \centering
        \includegraphics[width=\textwidth]{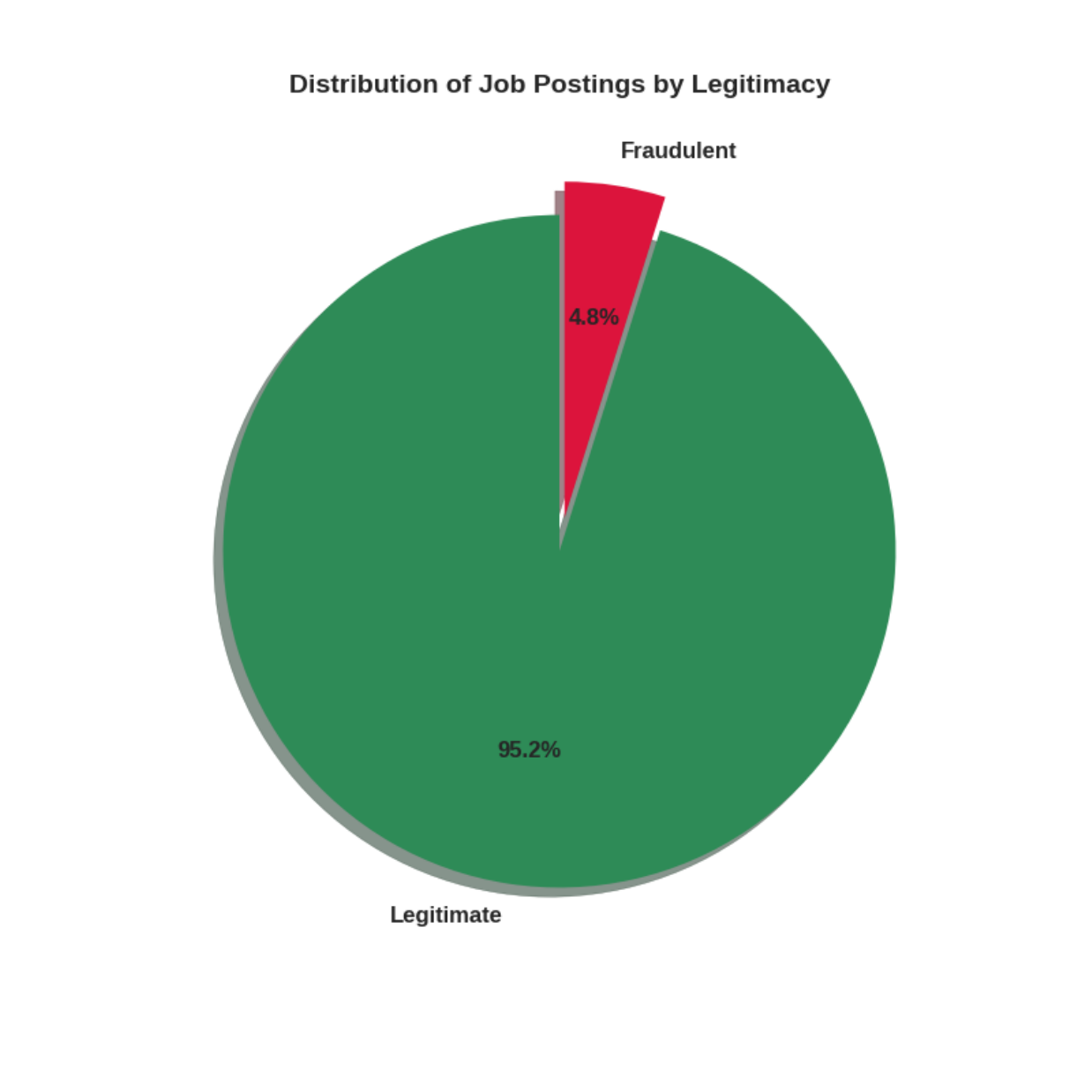}
    \end{minipage}%
    \begin{minipage}{0.58\textwidth}
        \centering
        \includegraphics[width=\textwidth]{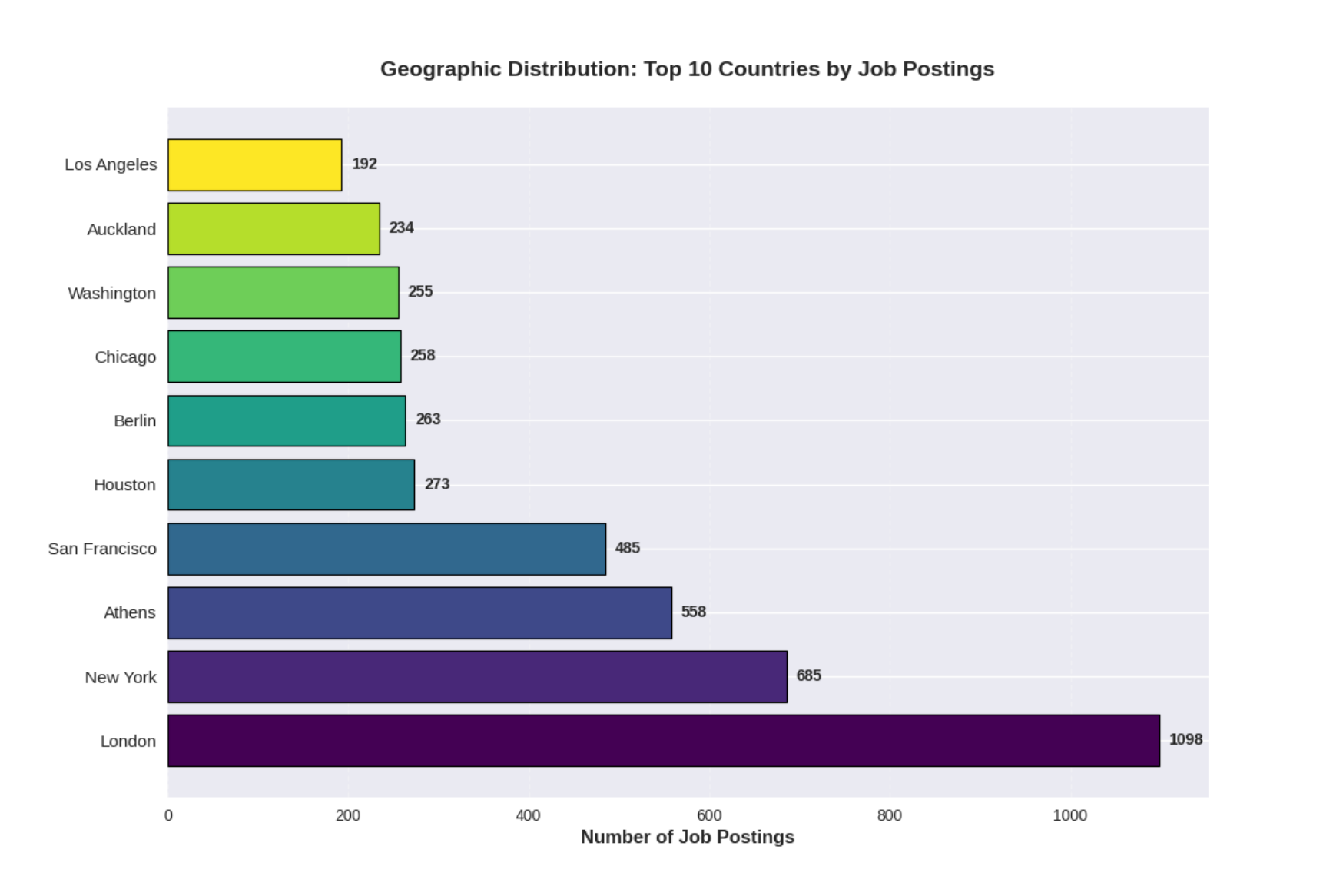}
    \end{minipage}
    
    \vspace{0.5em} % Add some vertical space between rows if desired
    
    \begin{minipage}{0.5\textwidth}
        \centering
        \includegraphics[width=\textwidth]{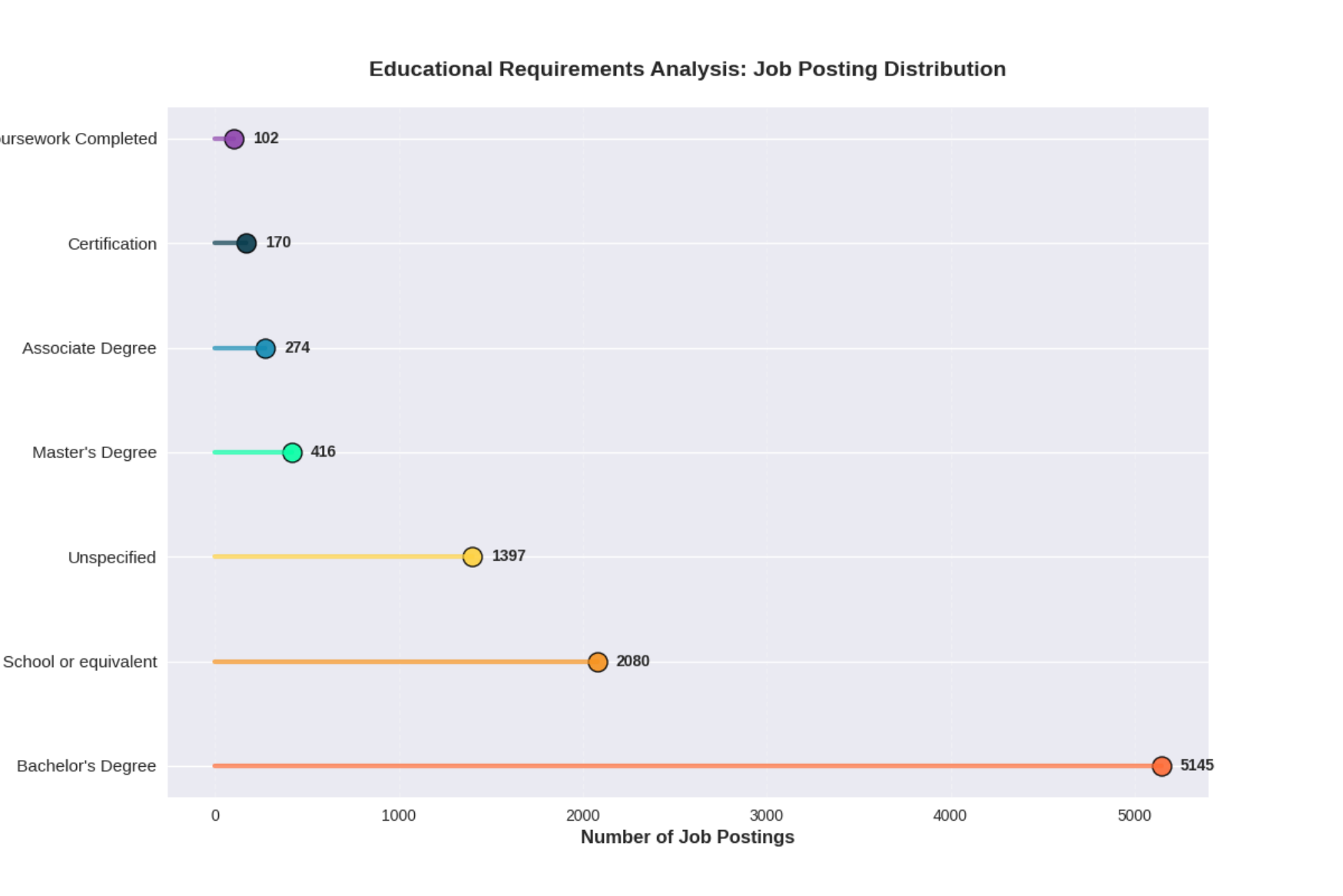}
    \end{minipage}%
    \begin{minipage}{0.58\textwidth}
        \centering
        \includegraphics[width=\textwidth]{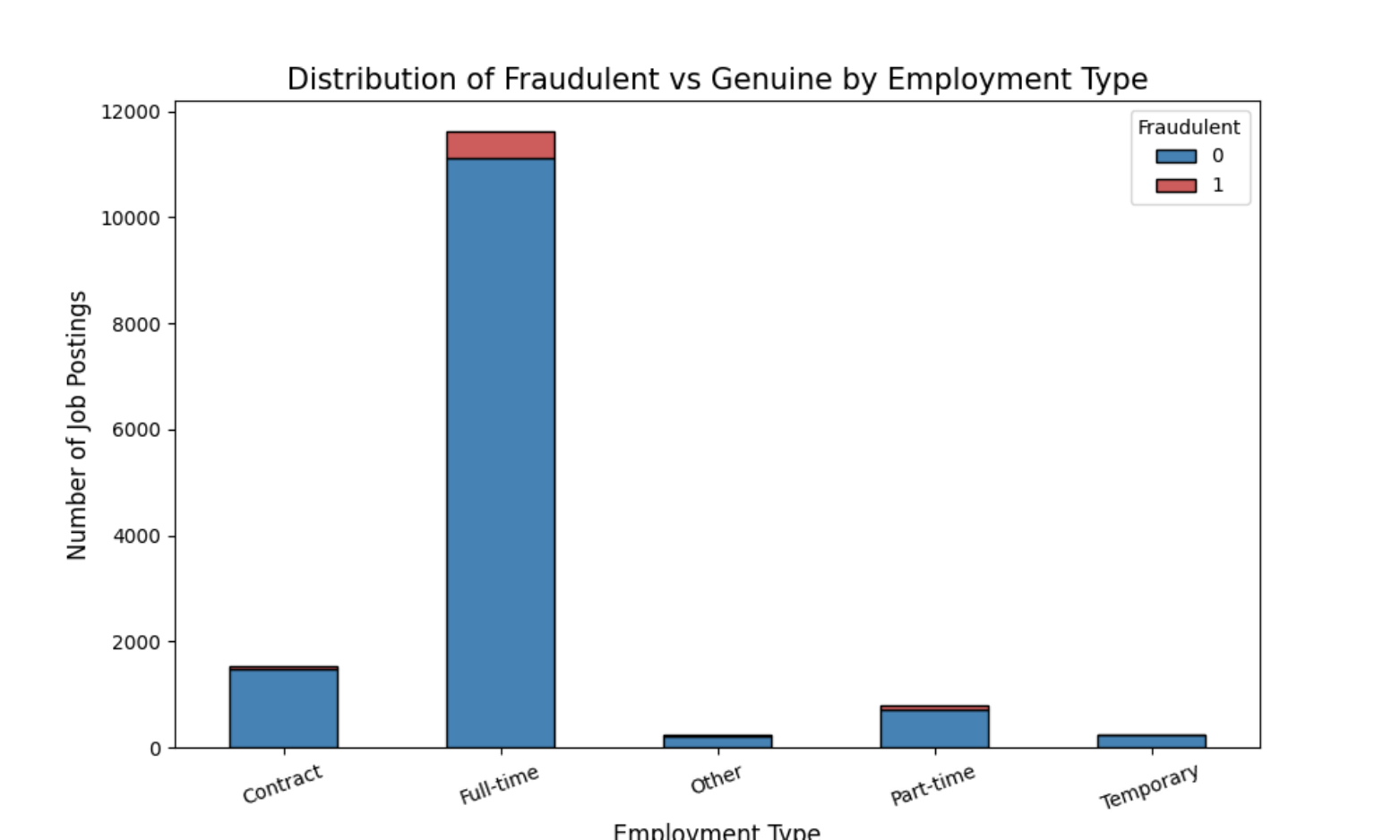}
    \end{minipage}
    \caption{Analysis of job postings: (a) Distribution of fraudulent vs. genuine postings, (b) Geographic distribution of postings, (c) Educational requirement analysis, and (d) Distribution by employment type}
    \label{fig:combined_analysis}
\end{figure*}

\section{\MakeUppercase{Related Work}}

In this section, we review some of the novel works previously done by seasoned researchers that have contributed to advancing fake job post detection. These studies will serve as the context for our current approach and highlight key trends, challenges, and gaps that our work aims to address.

Researchers have extensively explored classical machine learning approaches for identifying fraudulent job listings. Anita \textit{et al}. in \cite{anita2021fake} employ several machine learning models to detect fake job posts. The data cleaning process is specifically emphasized as a critical step in their pipeline. Among the various models evaluated, Bidirectional LSTMs achieved the best results, reporting an accuracy of 98\%. Similarly, Anbarasu \textit{et al}. \cite{anbarasu2024fake} trained multiple machine learning algorithms, including Naive Bayes Classifier (NBC), Decision Tree Classifier (DTC), Multilayer Perceptron (MLP), and Stochastic Gradient Descent (SGD), with SGD yielding the best results, achieving an impressive 98.6\% overall accuracy. Another similar work evaluates a broad spectrum of data mining and classification algorithms for predicting fake job posts. Singh \textit{et al}. in \cite{singh2023fake} developed a fraud detection model using existing ML models on the EMSCAD dataset, achieving 97.4\% accuracy in identifying fake job postings. Their approach involved data preprocessing, feature selection, and ensemble classification.

TF-IDF is a widely used feature-extraction technique that transforms text into weighted vectors based on token importance. Keerthana \textit{et al}. \cite{keerthana2021accurate} applied a TF-IDF vectorizer during preprocessing to encode tabular data, which was then fed into various ML models, with the MLP classifier achieving the highest accuracy of 71\%. Similarly, Dutta \textit{et al}. \cite{dutta2020fake}, after applying necessary data preprocessing steps, fed the extracted features to several ML models, with the Random Forest Classifier outperforming others in terms of accuracy. In a novel work that explores ensemble-based learning for fake job detection, Shibly \textit{et al}.  \cite{shibly2021performance} leverage boosted decision trees and two-class decision forest algorithms to detect fake job postings. In the first algorithm, decision trees are arranged in an ensemble manner, where the errors of previous trees are corrected by subsequent ones before arriving at the final prediction. Two class decision forests, on the other hand, rely on aggregated outcomes from grouped decision trees.

Beyond classical methods, several studies have shifted towards deep learning architectures to better capture semantic patterns in job descriptions. In a detailed approach presented by Pillai \textit{et al}. in \cite{pillai2023detecting}, they propose a training framework built on several BiLSTM layers. First the numeric and textual data is converted into fixed-size numerical vectors using separate tokenization layers, followed by an embedding layer, multiple BiLSTM layers, and a merging mechanism that fuses textual and numerical features for downstream classification. Rathudi \textit{et al}. in \cite{rathudi2023fake} leverage bidirectional LSTMs with Word2Vec \cite{mikolov2013efficient}, a vectorizer that converts textual tokens into low dimensional vector representations based on their semantic context. Coupling these two approaches allows the framework to capture both the semantic meaning of individual words and the sequential dependencies within the text, resulting in an overall accuracy of 97.1\%.

Some studies have also emphasized the importance of data imbalance mitigation and feature selection. Amaar \textit{et al}. in \cite{amaar2022detection} employ both TF-IDF vectorizer and BoW (Bag of Words) techniques for feature extraction. The extracted features are then fed into six machine learning models to evaluate the best performing combination. Additionally, they achieve over 99\% accuracy by incorporating oversampling in their framework. In a distinct study, Afzal \textit{et al}. \cite{afzal2024identifying} argue the reason that existing works often fell short is because feature selection and class imbalance are mostly overlooked. They leverage Chi-Square and PCA techniques to select the most relevant features, and apply SMOTE for minority class oversampling to effectively address the issue of class imbalance. While the above studies rely primarily on traditional feature engineering methods, a recent work explored deep contextual representations, Qayyum \textit{et al}. \cite{qayyum2023frd} propose a novel feature extraction technique called Deep Contextualized Word Representation (DCWR), that employs a two-layered BiLSTM to generate context-aware encoded representation of words by modeling the likelihood of word sequences in both forward and backward directions. Furthermore, they apply PCA as a feature reduction technique to select the principal components that capture maximum variance in the data.

\section{\MakeUppercase{Methodology}}

\subsection{\MakeUppercase{Data Collection}}
\label{sec:data_collection}

To carry out experiments on our proposed approach, we employed The Employment Scam Aegean Dataset (EMSCAD) \cite{shivamb_fake_job_posting}, a publicly available benchmark dataset for detection of fraudulent job postings. The dataset is highly imbalanced, comprising 17,014 genuine job advertisements and 866 fraudulent ones, collected between 2012 and 2014. The tabular dataset contains both numerical and categorical columns, providing sufficient information to train complex neural networks capable of modeling intricate relationships between features. Table~\ref{tab:job_dataset_schema} presents description of each feature along with their datatypes.

 \subsection{\MakeUppercase{Data Analysis}}
Data Analysis is a crucial step in building a machine learning model as it offers comprehensive insights related to our underlying dataset and aids in determining the selection of an appropriate modeling workflow. We first examined the distribution of target label (Figure~\ref{fig:combined_analysis}) (a), which reveals a severe class imbalance, with legitimate postings significantly exceeding fraudulent ones. The handling of such a problem is of utmost importance as it can bias the model towards underfitting and can lead to poor generalization over unseen examples of fraudulent postings. The solution to this problem shall later be discussed in the paper.

Additionally, analyzing geographical aspect of the dataset disclosed that majority of the advertisements came from cities like London, New York, and Athens that are considered the employment hubs of their respective countries as shown in Figure~\ref{fig:combined_analysis} (b). This shows us that the distribution of postings is not random, but is sophistically linked to the regions that are employment-concentrated. We also investigated the educational qualifications required by the postings (Figure~\ref{fig:combined_analysis}) (c), which ranged from school-level or equivalent to bachelor’s and master’s degrees, with the majority of jobs demanding at least an undergraduate degree, reflecting the diversity in posted advertisements across employment sectors.

Furthermore, we inspect what kind of employment types are more prone to get advertised in fraudulent postings, which revealed that \textbf{Full-time} positions exhibit a higher share of fraudulent advertisements compared to other positions as shown in Figure~\ref{fig:combined_analysis} (d). Finally, the data analysis provides a clear picture of all the challenges that need to be resolved before proceeding with effective preprocessing and model development.

\subsection{\MakeUppercase{Data Preprocessing}}

\subsubsection{\MakeUppercase{Data Cleaning}}
We first cleaned our data through a preprocessing pipeline. Data cleaning involves addressing inaccuracies, missing entries, duplicates, and inconsistently formatted data. We began by handling null values in the data which posed a significant challenge as there were a total of 70183 void entries in the dataset. However, this number is aggregated across all columns and therefore can be misleading. To obtain a more accurate assessment, we only selected the columns that contained some amount of null values, and computed the average number of nulls per affected column which resulted in approximately 5848 null entries per column which still is a very concerning number. To handle this, we removed the integer columns from our dataset, retaining only the categorical features that were sufficient for modeling complex patterns. Following this, all the null entries were then replaced with empty strings. 

These empty strings individually seem to appear meaningless and insignificant, however, when concatenated with other categorical columns, we obtain a combined column named "text". This methodology preserves the structure of the data, as it eliminates the need to remove rows with missing values, thereby maintaining the overall size and integrity of the dataset. 

Furthermore, we removed all the duplicated rows from our dataset to maintain consistency, and performed additional pre-processing on the remaining textual data. In order to do this, we designed a pipeline that carried out several cleaning tasks including converting text to lowercase, removing emails, URLs, and HTML tags, eliminating punctuation and numbers, removing stopwords, and finally lemmatizing the tokens using spaCy, ensuring that our data is suitable for feature engineering and machine learning modeling.

\subsubsection{\MakeUppercase{Feature Extraction}}
Once the dataset was structurally cleaned, we employed several feature extraction methods such as \textit{TF-IDF Vectorizer, Word2Vec}, and \textit{Glove} to obtain feature embeddings. Among these, the method that we finally adopted was Glove as it showed improved and consistent results with our clustering approach. Since Glove is a well-established embedding technique, that captures global context by estimating the co-occurance probability between two words, we were able to obtain \(n^{\text{th}}\)-dimensional embeddings (where \(n=300\)) for each word in the vocabulary with minimal parameter tuning. Therefore, we focus less on its implementation details here and instead present its comparative performance in Section~\ref{sec:results}.

\begin{algorithm}
    \caption{Centroid-Guided Contrastive Loss (CGCL)}
    \label{alg:cgcl}
    \KwIn{Embeddings $E$, Logits $Z$, Labels $Y$, 
    hyperparameters: $k$, $\alpha$, $\beta$, margin $m$}
    \KwOut{Loss $\mathcal{L}$}
    
    \BlankLine
    \textbf{Step 1: Classification Loss (Weighted CE)} \\
    Compute class counts: $n_c = \text{count}(Y=c)$ for each class $c$ \\
    Compute weights: $w_c = \frac{1}{n_c + \epsilon}$ \\
    $\mathcal{L}_{CE} \gets - \frac{1}{N} \sum_{i=1}^N w_{y_i} \log p(y_i \mid x_i)$
    
    \BlankLine
    \textbf{Step 2: Normalize Embeddings} \\
    $E \gets \frac{E}{\|E\|_2}$
    
    \BlankLine
    \textbf{Step 3: Contrastive Loss (Centroid-Guided)} \\
    Initialize $\mathcal{L}_{contrastive} \gets 0$, $C \gets 0$  \\
    \ForEach{class $c \in \text{unique}(Y)$}{
        $E_c \gets \{ e_i \in E \mid y_i = c \}$ \\
        $E_{\neg c} \gets \{ e_i \in E \mid y_i \neq c \}$ \\
        \If{$|E_c| < k+1$ or $|E_{\neg c}| < k$}{
            \textbf{continue}
        }
        Compute centroid: $\mu_c = \frac{1}{|E_c|}\sum_{e \in E_c} e$
    
        \BlankLine
        \tcp{Pull: hardest positives}
        $d^{+} = \{ \|e - \mu_c\|_2 \mid e \in E_c \}$ \\
        Select top-$k$ farthest: $H^{+}_k$ \\
        $\mathcal{L}_{pull} = \frac{1}{k}\sum_{e \in H^{+}_k} \|e - \mu_c\|_2^2$
    
        \BlankLine
        \tcp{Push: hardest negatives}
        $d^{-} = \{ \|e - \mu_c\|_2 \mid e \in E_{\neg c} \}$ \\
        Select top-$k$ closest: $H^{-}_k$ \\
        $\mathcal{L}_{push} = \frac{1}{k}\sum_{e \in H^{-}_k} \max(0, m - \|e - \mu_c\|_2)^2$
    
        \BlankLine
        \tcp{Combine push-pull}
        $\mathcal{L}_{contrastive} \gets \mathcal{L}_{contrastive} + (1-\alpha)\mathcal{L}_{pull} + \alpha \mathcal{L}_{push}$ \\
        $C \gets C + 1$
    }
    \If{$C > 0$}{
        $\mathcal{L}_{contrastive} \gets \mathcal{L}_{contrastive} / C$
    }
    
    \BlankLine
    \textbf{Step 4: Unified Loss} \\
    $\mathcal{L} \gets \beta \mathcal{L}_{CE} + (1-\beta)\mathcal{L}_{contrastive}$ \\
    \Return{$\mathcal{L}$}
    \end{algorithm}

\begin{figure*}[!t]
\centering
\includegraphics[width=0.55\textwidth, keepaspectratio]{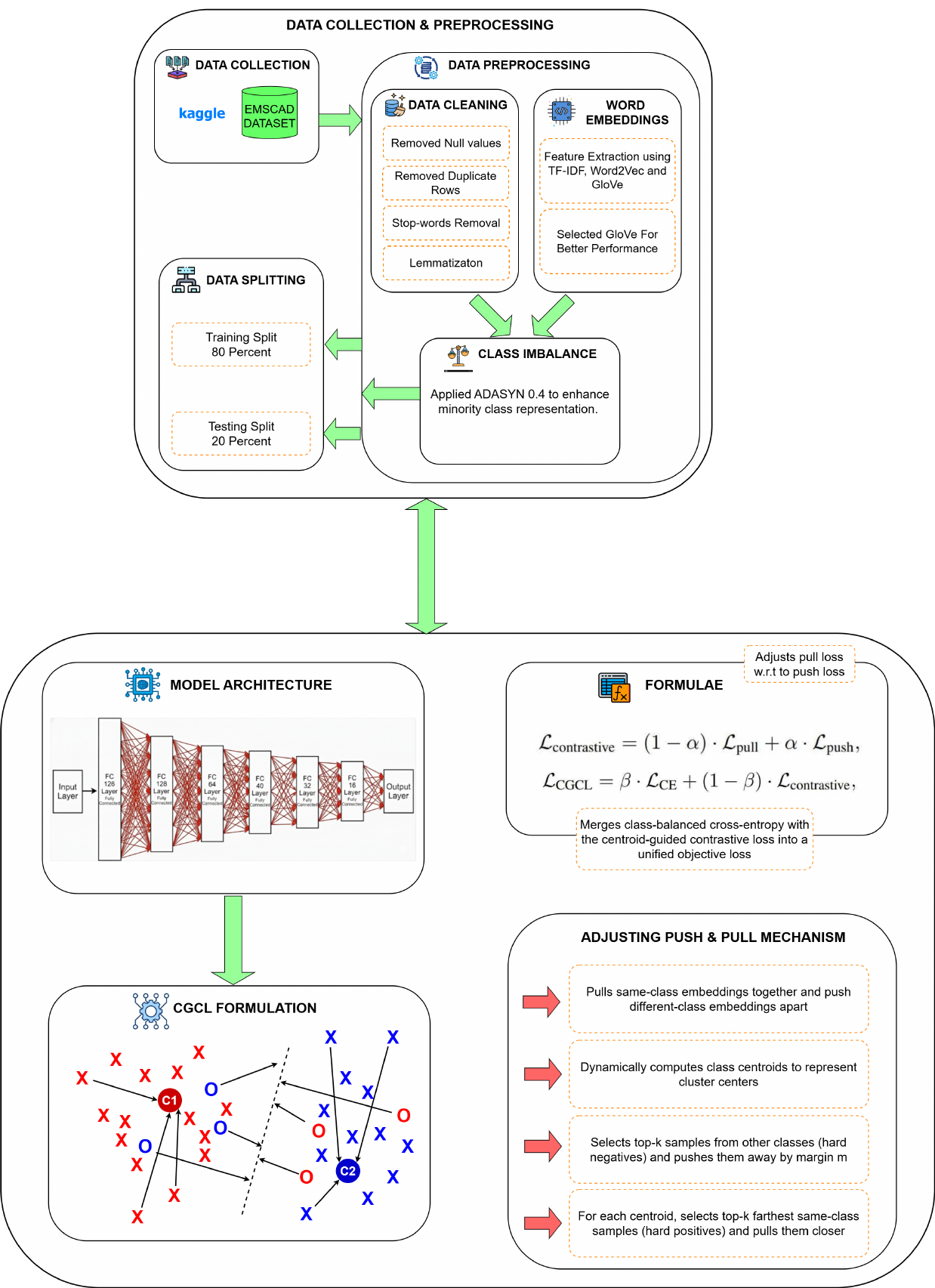}
\caption{The pipeline consists of two phases: (1) Data Preprocessing \& Augmentation, where raw text from the EMSCAD dataset
is cleaned, vectorized using Word2Vec, and balanced using the ADASYN oversampling technique. (2) Deep Metric Learning,
featuring a hierarchical Feed-Forward Neural Network (Input to FC-16) optimized by the proposed Centroid-based Geometric
Contrastive Loss (CGCL). The CGCL formulation (bottom center) minimizes intra-class variance by pulling samples toward
their respective centroids (C1, C2) while maximizing inter-class separability.}
\label{fig:fig4}
\end{figure*}

\subsubsection{\MakeUppercase{Imbalance Handling}}

Imbalance handling is a necessary step when the dataset is largely inclined towards the distribution of a certain class samples. Class imbalance hinders the generalization capability of machine learning models as they get biased towards majority class samples leading to the model overfitting.

To handle class imbalance, we leveraged two complementary strategies, first, at the data level, we applied \textit{ADASYN (Adaptive Synthetic Sampling)} to generate synthetic minority samples with a sampling ratio of 0.4, enabling the model to capture intricate representations of minority class for reliable classification. The choice of ADASYN over \textit{SMOTE (Synthetic Minority Oversampling Technique)} was because ADASYN oversamples the minority class by focusing on its local context, thereby generating minority class samples that are harder to classify, whereas SMOTE produces uniformly distributed synthetic samples that may not sufficiently capture minority complexity.

Second, at algorithmic level, we used a class-weighted cross-entropy loss, where we assigned weights to the class based on their frequencies. The class having majority samples was given a lower weight while the one having lesser samples was allocated a higher weight, ensuring that misclassifications of the minority class were penalized more heavily during training.

\subsection{\MakeUppercase{Loss Formulation}}

    In this section, we propose \textit{CGCL (Centroid-Guided Contrastive Loss)}, a custom loss function that unifies both classification and clustering within a single optimization objective. While many existing works treat fraudulent postings detection as merely a classification task, we argue that relying solely on \textit{Cross-Entropy} loss is suboptimal, as it does not modulate the structure of latent-embedding space explicitly. Additionally, leveraging only the classic \textit{Contrastive} loss would be ineffective due to its uniform pairwise distance computations, which can dilute the focus on harder examples. CGCL on the other hand, addresses these limitations by incorporating a classification loss that enforces discriminative decision boundaries and a centroid-driven contrastive loss to reshape the embedding space, ensuring intra-class compactness and inter-class separability.

    \subsubsection{\MakeUppercase{Weighted Cross-Entropy Component}}
    To handle class-imbalance, we adopt a weighted variant of the standard Cross-Entropy loss. We do this by assigning weights to the classes in inverse proportion to their frequency in the dataset, unlike the vanilla CE loss which treats all classes equally. This approach gives more importance to the minority class while significantly reducing the dominance of majority classes. Formally, the loss is given by:

    \begin{equation}
    \label{eq:cross}
    \mathcal{L}_{CE} = - \frac{1}{N} \sum_{i=1}^{N} w_{y_i} \, \log \big( p(y_i \mid x_i) \big),
    \end{equation}

    where $N$ is the total number of samples, 
    $p(y_i \mid x_i)$ is the predicted probability of the ground-truth class $y_i$ for input $x_i$, 
    and $w_{y_i}$ denotes the class weight.

    \vspace{1em}

    \subsubsection{\MakeUppercase{Centroid-Guided Push and Pull Mechanism}}

    This is the core of our proposed algorithm. Taking inspiration from the well-known Contrastive loss \cite{khosla2020supervised}, we encourage embeddings of samples from the same class to be pulled closer together, while embeddings of samples from different classes are pushed apart in the latent space. However, unlike traditional contrastive loss, where pairwise distances are computed exhaustively between samples, we design a novel centroid-driven strategy where centroids for unique classes are computed dynamically to represent their respective clusters. For every centroid, we identify the farthest top-$k$ samples (hard-positives) belonging to the same class as the active centroid, and pull them closer to the centroid \eqref{eq:pull}, thereby reinforcing intra-class compactness. Similarly, we locate top-$k$ from other classes (hard-negatives) and push them away from the active centroid by a margin \(m\) \eqref{eq:push}, which can be adjusted during training, ensuring stronger inter-class separability. We then combine this push-pull mechanism in a unified framework, yielding a novel variant of contrastive loss, as formulated in \eqref{eq:contrastive}.

    \begin{equation}
    \label{eq:pull}
    \mathcal{L}_{\text{pull}} = \frac{1}{K} \sum_{k=1}^{K} \left\| f(x^{\text{hard}}_{k}) - \mu_c \right\|^2_2,
    \end{equation}
    
    where \(x^{\text{hard}}_{k}\) represents the \(k\)-th hardest positive sample (farthest from the centroid \(\mu_{c}\)).

    \begin{equation}
    \label{eq:push}
    \mathcal{L}_{\text{push}} = \frac{1}{K} \sum_{k=1}^{K} \max \Big(0, \, m - \| f(x^{\text{neg}}_{k}) - \mu_c \|_2 \Big)^2,
    \end{equation}

    where \(x^{\text{neg}}_{k}\) represents the hard negatives and \(m\) is the margin parameter.

    \begin{equation}
    \mathcal{L}_{\text{contrastive}} 
    = (1 - \alpha) \cdot \mathcal{L}_{\text{pull}} 
    + \alpha \cdot \mathcal{L}_{\text{push}},
    \label{eq:contrastive}
    \end{equation}

    where $\alpha \in [0,1]$ balances the contribution of push loss with respect to the pull loss.

    \vspace{1em}
    \subsubsection{\MakeUppercase{Unified Loss Function}}

    Finally, we integrate the class-balanced Cross-Entropy component \eqref{eq:cross} with the centroid-guided contrastive objective \eqref{eq:contrastive} into a unified loss formulation. This combined loss not only enforces robust class separation through discriminative decision boundaries but also explicitly reshapes the latent embedding space via the push-pull mechanism. Formally, the complete loss is given by:
    
     \begin{equation}
    \mathcal{L}_{\text{CGCL}} 
    = \beta \cdot \mathcal{L}_{\text{CE}} 
    + (1 - \beta) \cdot \mathcal{L}_{\text{contrastive}},
    \label{eq:cgcl}
    \end{equation}
    
    where $\beta \in [0,1]$ controls the balance between the classification term and the contrastive objective. Moreover, all embeddings are $\ell_{2}$-normalized to maintain consistent scale across samples. For clarity, the step-by-step procedure of the proposed CGCL is summarized in Algorithm~\ref{alg:cgcl}.

    \subsection{\MakeUppercase{Model Architecture}}

For performing classification, we adopt a straightforward Multi-layer Perceptron (MLP) architecture. First the embeddings are obtained via Glove, which are then passed through the feed-forward neural network. The model consists of six fully connected layers, each followed by a ReLU activation function which enables the model to capture non-linearity in the data. These deep layers significantly reduce the dimensionality of input features from 300-dimensional GloVe embeddings to a compact latent representation of size 16. The resultant embeddings are then sent into a linear layer for classification, formally the model can be expressed as:

\begin{equation}
    z = f_{\text{embed}}(x)
\end{equation}

\begin{equation}
    \text{logits}  = W_{\text{cls}} z + b_{\text{cls}}
\end{equation}

$z$ is the latent representation, 
$W_{\text{cls}}$ are the classifier weights, and 
$b_{\text{cls}}$ is the classifier bias.

    \begin{figure}[htbp]
    \centering
    \begin{tikzpicture}[
        node distance=1cm,
        % Base node style
        base/.style={
            draw,
            rounded corners=8pt,
            minimum width=3.5cm,
            minimum height=1cm,
            align=center,
            font=\bfseries,
            drop shadow={
                shadow xshift=3pt,
                shadow yshift=-3pt,
                fill=shadowcolor!40
            }
        },
        % Input layer style
        input/.style={
            base,
            fill=inputcolor!20,
            draw=inputcolor!80,
            line width=2pt
        },
        % Hidden layer style
        hidden/.style={
            base,
            fill=hiddencolor!20,
            draw=hiddencolor!80,
            line width=2pt
        },
        % Embedding layer style
        embed/.style={
            base,
            fill=embedcolor!20,
            draw=embedcolor!80,
            line width=2pt
        },
        % Output layer style
        output/.style={
            base,
            fill=outputcolor!20,
            draw=outputcolor!80,
            line width=2pt
        },
        % Arrow style
        arrow/.style={
            -Stealth,
            line width=3pt,
            color=arrowcolor
        },
        % Dimension labels
        dimstyle/.style={
            font=\small\itshape,
            color=arrowcolor!70
        }
    ]
    
    % Input layer
    \node[input] (input) {
        \Large Input Layer \\
        \normalsize 300-dim GloVe \\
        \footnotesize Word Embeddings
    };
    
    % Hidden layers with progressive styling
    \node[hidden, below=of input] (fc1) {
        \Large Hidden Layer 1 \\
        \normalsize FC: 256 + ReLU \\
        \footnotesize Feature Extraction
    };
    
    \node[hidden, below=of fc1] (fc2) {
        \Large Hidden Layer 2 \\
        \normalsize FC: 128 + ReLU \\
        \footnotesize Dimensionality Reduction
    };
    
    \node[hidden, below=of fc2] (fc3) {
        \Large Hidden Layer 3 \\
        \normalsize FC: 64 + ReLU \\
        \footnotesize Feature Compression
    };
    
    \node[hidden, below=of fc3] (fc4) {
        \Large Hidden Layer 4 \\
        \normalsize FC: 40 + ReLU \\
        \footnotesize Abstraction Layer
    };
    
    \node[hidden, below=of fc4] (fc5) {
        \Large Hidden Layer 5 \\
        \normalsize FC: 32 + ReLU \\
        \footnotesize Semantic Encoding
    };
    
    % Embedding layer (special styling)
    \node[embed, below=of fc5] (fc6) {
        \Large Embedding Layer \\
        \normalsize FC: 16 \\
        \footnotesize \textit{Latent Representation}
    };
    
    % Output layer
    \node[output, below=of fc6] (output) {
        \Large Classifier \\
        \normalsize 2 Classes \\
        \footnotesize Binary Classification
    };
    
    % Enhanced arrows with labels
    \draw[arrow] (input) -- (fc1) node[midway, right=0.3cm, dimstyle] {$\mathbb{R}^{300}$};
    \draw[arrow] (fc1) -- (fc2) node[midway, right=0.3, dimstyle] {$\mathbb{R}^{256}$};
    \draw[arrow] (fc2) -- (fc3) node[midway, right=0.3, dimstyle] {$\mathbb{R}^{128}$};
    \draw[arrow] (fc3) -- (fc4) node[midway, right=0.3, dimstyle] {$\mathbb{R}^{64}$};
    \draw[arrow] (fc4) -- (fc5) node[midway, right=0.3, dimstyle] {$\mathbb{R}^{40}$};
    \draw[arrow] (fc5) -- (fc6) node[midway, right=0.3, dimstyle] {$\mathbb{R}^{32}$};
    \draw[arrow] (fc6) -- (output) node[midway, right=0.3, dimstyle] {$\mathbb{R}^{16}$};

    \end{tikzpicture}
    
    \caption{Enhanced MLP Embedder Architecture. This diagram illustrates a seven-layer fully connected neural network used for feature extraction and latent representation learning.}

    \label{fig:mlp_architecture}
\end{figure}
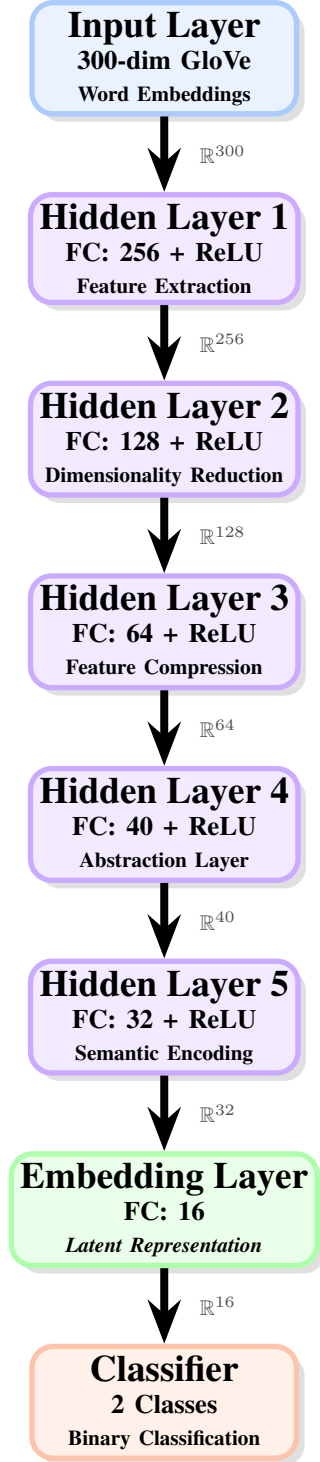

\section{\MakeUppercase{Experimental Results}}
\label{sec:results}
We validate our proposed approach on a benchmark dataset in comparison with well-known studies previously done in the field of fraudulent posts detection. We perform extensive experiments to justify the effectiveness of our method, including model-level
assessments through hyperparameter tuning and training configurations, and data-level evaluations to analyze the influence of various feature extractors such as GloVe, Word2Vec, and TF–IDF. on the performance of our model.

\begin{table}[ht!]
\centering
\caption{\textbf{Performance of Word2Vec, GloVe, and TF-IDF embeddings under identical hyperparameters} 
($\textbf{$\alpha$}=0.6, \textbf{$\beta$}=0.9, k=3, m=1.0$). Values are reported as \textit{Macro / Micro}. 
Best results are in \textbf{bold}.}
\label{tab:base_results}
\renewcommand{\arraystretch}{0.95}
\setlength{\tabcolsep}{3pt}
\begin{tabular}{llcccc}
\toprule
\textbf{Embedding} & \textbf{Epochs} & \textbf{Accuracy} & \textbf{Precision} & \textbf{Recall} & \textbf{F1-Score} \\
\midrule\midrule
\multirow{3}{*}{Word2Vec} 
 & 1500 & 0.981 & 0.97 / 0.98 & 0.98 / 0.98 & 0.98 / 0.98 \\
 & 2000 & 0.985 & 0.98 / 0.99 & 0.99 / 0.99 & 0.98 / 0.99 \\
 & 3000 & \textbf{0.986 } & \textbf{0.98 / 0.99} & \textbf{0.99 / 0.99} & \textbf{0.98 / 0.99} \\
\midrule
\multirow{3}{*}{GloVe} 
 & 1500 & 0.985 & 0.98 / 0.99 & 0.99 / 0.99 & 0.98 / 0.98 \\
 & 2000 & 0.981 & 0.97 / 0.98 & 0.98 / 0.98 & 0.98 / 0.98 \\
 & 3000 & \textbf{0.990} & \textbf{0.98 / 0.99} & \textbf{0.99 / 0.99} & \textbf{0.98 / 0.99} \\
\midrule
\multirow{2}{*}{TF-IDF} 
 & 500  & 0.989 & 0.98 / 0.99 & 0.99 / 0.99 & 0.98 / 0.99 \\
 & 1000 & \textbf{0.991} & \textbf{0.99 / 0.99} & \textbf{0.99 / 0.99} & \textbf{0.99 / 0.99} \\
\bottomrule
\end{tabular}
\end{table}

\sisetup{detect-weight=true, detect-family=true}

\begin{table}[ht]
\centering
\caption{Top hyperparameter configurations for GloVe and Word2Vec embeddings.}
\label{tab:hyper_results_combined}
\begin{tabular}{lccc
    S[table-format=1.3, detect-weight, detect-family]
    S[table-format=1.3, detect-weight, detect-family]}
\toprule
\textbf{Embedding} & \textbf{$\alpha$} & \textbf{$\beta$} & \textbf{$K$} &
\textbf{Accuracy} & \textbf{F1} \\
\midrule
\multirow{5}{*}{GloVe}
& 0.5 & 0.7 & 5  & \bfseries 0.992 & \bfseries 0.987 \\
& 0.5 & 0.5 & 10 & 0.987 & 0.978 \\
& 0.5 & 0.5 & 7  & 0.986 & 0.977 \\
& 1.0 & 0.3 & 10 & 0.986 & 0.976 \\
& 1.0 & 0.5 & 7  & 0.984 & 0.973 \\
\midrule
\multirow{5}{*}{Word2Vec}
& 0.5 & 0.3 & 7  & \bfseries 0.989 & \bfseries 0.981 \\
& 1.0 & 0.5 & 5  & 0.988 & 0.980 \\
& 1.0 & 0.7 & 7  & 0.988 & 0.979 \\
& 1.0 & 0.7 & 5  & 0.987 & 0.978 \\
& 0.5 & 0.3 & 10 & 0.986 & 0.976 \\
\bottomrule
\end{tabular}
\label{tab:hyp}
\end{table}

\vspace{1em}
\subsection{\MakeUppercase{Implementation Details}}

Every experiment was carried out in PyTorch on a system with 64GB of RAM and an NVIDIA RTX 4090 GPU. We assessed three representations for feature extraction: TF-IDF, Word2Vec, and GloVe (300-dimensional pretrained embeddings). The proposed MLP embedder was trained using the Adam optimizer with an initial learning rate of $1 \times 10^{-3}$, batch size of 128, and early stopping based on validation loss. The training was carried out across various epochs such as 500, 1500, 2000 and 3000. For our CGCL loss, we explored different settings of the hyperparameters $\alpha$, $\beta$, $k$, and $m$, while keeping all embeddings $\ell_{2}$-normalized. The dataset was split into 80\% training and 20\% testing, with results reported on the held-out test set.

\begin{table}[ht]
\centering
\caption{\textbf{Best performing models for each embedding method.} 
Listed are the optimal hyperparameters and corresponding performance scores.}
\label{tab:best_models}
\renewcommand{\arraystretch}{1.05}
\setlength{\tabcolsep}{8pt}
\begin{tabular}{lccccc}
\hline
\textbf{Embedding} & $\boldsymbol{\alpha}$ & $\boldsymbol{\beta}$ & $\boldsymbol{k}$ & \textbf{Accuracy} & \textbf{F1-Score} \\
\hline\hline
Word2Vec & 0.5 & 0.3 & 7 & 0.989 & 0.981 \\
TF-IDF   & 0.6 & 0.9 & 3 & 0.991 & 0.990 \\
\hline
GloVe    & 0.5 & 0.7 & 5 & \textbf{0.992} & 0.987 \\
\hline
\end{tabular}
\end{table}

\begin{table}[ht]
\centering
\caption{\textbf{Clustering evaluation metrics for the final selected model (GloVe, $\alpha=0.5, \beta=0.7, k=5$).}}
\label{tab:clustering_metrics}
\renewcommand{\arraystretch}{1.05}
\setlength{\tabcolsep}{10pt}
\begin{tabular}{lc}
\hline
\textbf{Metric} & \textbf{Score} \\
\hline\hline
Silhouette Score          & 0.701 \\
Adjusted Rand Index (ARI) & 0.953 \\
Normalized Mutual Information (NMI) & 0.903 \\
Homogeneity Score         & 0.909 \\
Completeness Score        & 0.898 \\
V-Measure                 & 0.903 \\
Calinski-Harabasz Score   & 5248.366 \\
Davies-Bouldin Score      & 0.777 \\
\hline
\end{tabular}
\end{table}

\begin{figure*}[h]
    \centering
    \begin{minipage}{0.32\textwidth}
        \centering
        \includegraphics[width=\textwidth]{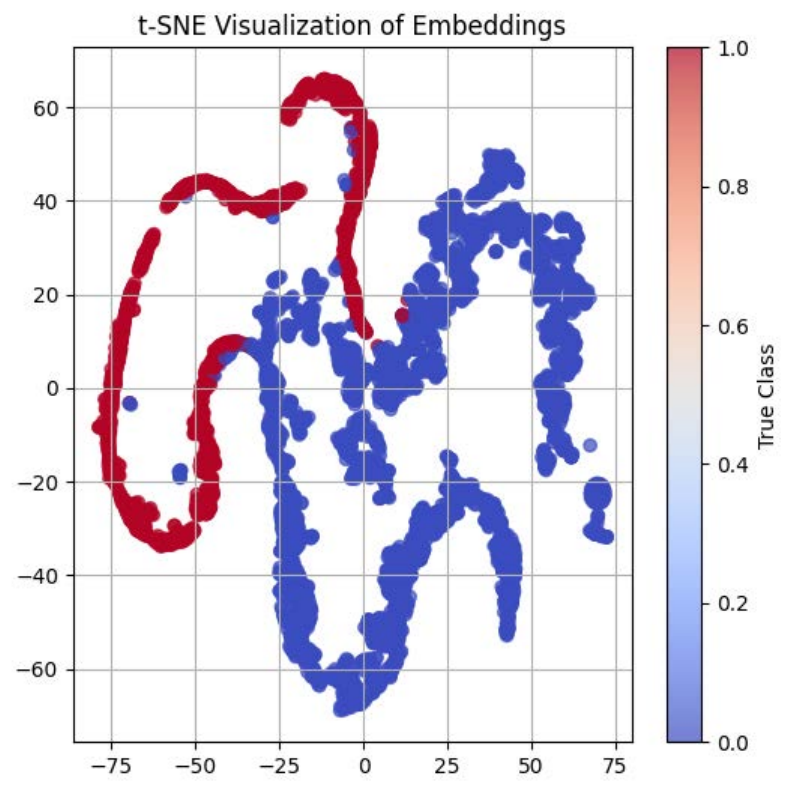}
    \end{minipage}
    \hspace{0.03\textwidth}
    \begin{minipage}{0.32\textwidth}
        \centering
        \includegraphics[width=\textwidth]{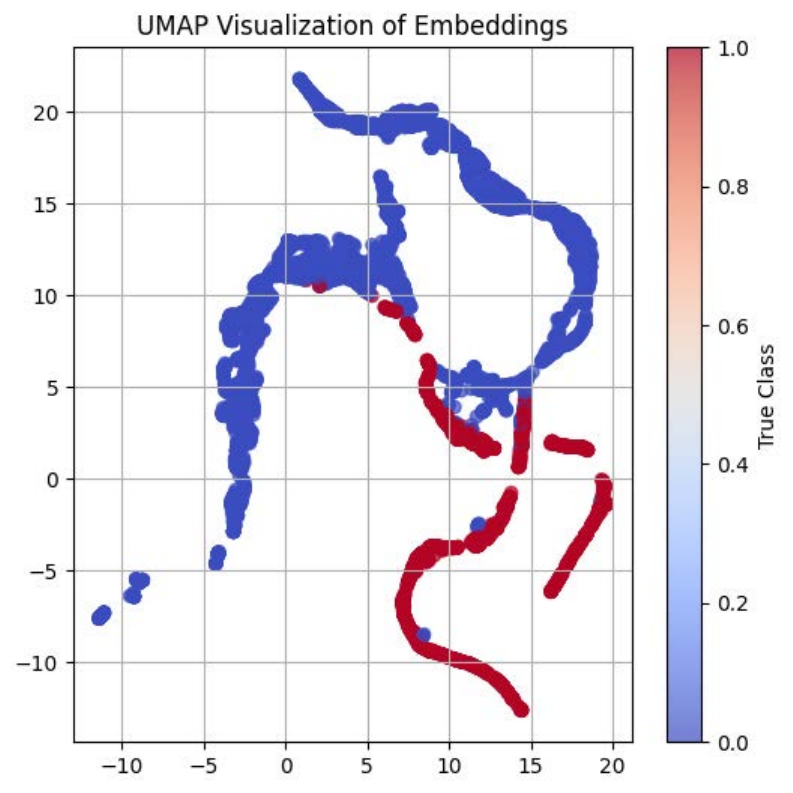}
    \end{minipage}
    \caption{t-SNE (left) reveals the continuous internal
    structure of fraud templates, while UMAP (right) confirms the global separability and distinct sub-clustering of legitimate job categories.}
    \label{fig:tsne_umap}
\end{figure*}

\subsection{\MakeUppercase{Results}}

The first set of experiments evaluated different feature extraction techniques under identical hyperparameters across multiple epochs (1500, 2000, and 3000 for Word2Vec and GloVe). For the TF-IDF vectorizer, experiments were conducted at 500 and 1000 epochs due to its faster convergence. Beyond 1000 epochs, no significant performance gains were observed with TF-IDF. To ensure a fair comparison, the hyperparameters were kept fixed across all experimental settings. Details are shown in Table~\ref{tab:base_results}.

\begin{figure}[htbp]
\centering
\includegraphics[width=\columnwidth]{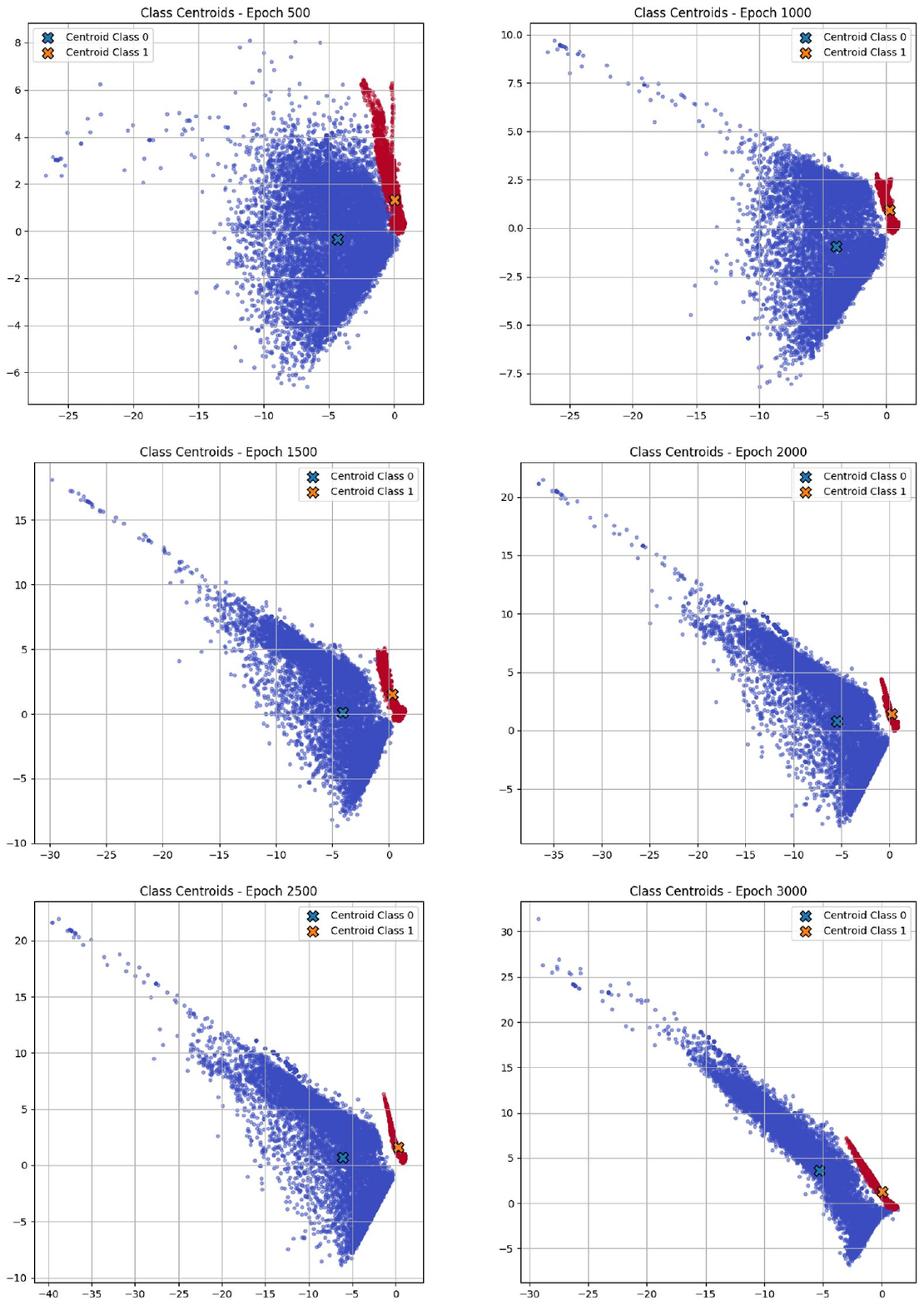}
\caption{Evolution of Latent Space. The progression from epoch
500 to 3000 shows the ’push’ mechanism creating a gradient
of confidence for legitimate jobs (blue) while compacting
fraudulent jobs (red) into a dense anomaly cluster.\\
Hyperparameters: ($\alpha = 0.5, \; \beta = 0.7, \; k = 5$)}
\label{fig:evolution}
\end{figure}

\begin{table*}[!t]
\centering
\caption{Test-set metrics (mean $\pm$ std, five seeds, fraud class). Best per column in \textbf{bold}.}
\label{tab:ablation_results}
\setlength{\tabcolsep}{5pt}
\renewcommand{\arraystretch}{1.25}
\small
\begin{tabular}{lC{1.8cm}C{1.8cm}C{1.8cm}C{1.8cm}C{1.8cm}C{1.8cm}}
\toprule
\textbf{Variant} & \textbf{Accuracy} & \textbf{Precision} & \textbf{Recall} & \textbf{F1} & \textbf{PR-AUC} & \textbf{MCC} \\
\midrule
CE\_only       & 0.974{\scriptsize±.002} & 0.763{\scriptsize±.036} & 0.677{\scriptsize±.015} & 0.718{\scriptsize±.021} & 0.727{\scriptsize±.022} & 0.706{\scriptsize±.023} \\
CE\_weighted   & 0.976{\scriptsize±.003} & 0.772{\scriptsize±.043} & 0.719{\scriptsize±.022} & 0.744{\scriptsize±.026} & 0.740{\scriptsize±.030} & 0.732{\scriptsize±.028} \\
CE\_ADASYN     & 0.989{\scriptsize±.000} & 0.967{\scriptsize±.002} & 0.995{\scriptsize±.003} & 0.981{\scriptsize±.001} & 0.982{\scriptsize±.003} & 0.973{\scriptsize±.001} \\
\rowcolor{bestrow}
CE\_CenterLoss & \textbf{0.990}{\scriptsize±.001} & \textbf{0.968}{\scriptsize±.005} & 0.997{\scriptsize±.002} & \textbf{0.982}{\scriptsize±.002} & 0.980{\scriptsize±.006} & \textbf{0.975}{\scriptsize±.003} \\
CGCL\_full     & 0.983{\scriptsize±.012} & 0.947{\scriptsize±.035} & \textbf{0.997}{\scriptsize±.002} & 0.971{\scriptsize±.020} & \textbf{0.981}{\scriptsize±.011} & 0.960{\scriptsize±.028} \\
\bottomrule
\end{tabular}
\end{table*}

Furthermore, since our designed loss function \textbf{CGCL} is highly sensitive to hyperparameter settings, in order to obtain the best performing model for each technique, we conducted several experiments with various combinations of parameters. In total, 27 possible combinations were evaluated per technique. For clarity and conciseness, we present only the top 5 results for GloVe and Word2Vec, summarized in structured form in Table~\ref{tab:hyp}. Across these experiments, we observed that $\alpha$ within the range of 0.5-1.0 exhibited consistent results, while $\alpha=1.5$ consistently failed across both techniques. Additionally, the $\beta$ parameter displayed technique-specific behavior: The top results were obtained when $\beta$ was set to 0.5 on GloVe-based embeddings, whereas Word2Vec performed better with $\beta=0.3$, showing that the optimal setting of $\beta$ is embedding-dependent.

Table~\ref{tab:best_models} presents the best-performing model for each technique, making it convenient to identify the GloVe-based embedding model as the final selected model for our study. It achieves an impressive \textbf{99.2\%} overall accuracy  and an F1-score of \textbf{98.7\%} on our test set.

Finally, since our loss function is inherently incomplete without the \textit{contrastive} component to enforce tight clustering in the latent space and thereby complement the classification objective, we therefore present the evaluations of our final model on established clustering metrics in Table~\ref{tab:clustering_metrics}.

In conclusion, our proposed approach shows astounding classification performance with GloVe-based model achieving an overall accuracy of \textbf{99.2\%} and an F1-score of \textbf{98.7\%}. In addition, the clustering performance metrics further validate our model's robustness in a complex classification task, thereby proving its representational and discriminative efficacy.

\subsection{\MakeUppercase{Visualizations and Latent Space Analysis}}

To prove the effectiveness and authority of CGCL and to understand how latent space evolves during training, we visualized the high-dimensional embeddings ($d$ = 16) projected into 2D space. These visualizations validate that the custom dual-objective loss function successfully enforces both intraclass compactness and inter-class separability.

\subsubsection{\MakeUppercase{Temporal Evolution of Class Separation and Inter-Class Compactness}}

Figure~\ref{fig:evolution} illustrates the evolution of the latent space over 3000 training epochs. The blue points represent Non-Fraudulent (Class 0) job postings, while the red points represent Fraudulent (Class 1) postings. At earlier epochs, there is substantial overlap between the two classes, particularly at epoch 500. As training progresses, the embeddings gradually organize into two distinct regions, and by epoch 3000 the separation becomes considerably clearer.

The legitimate job postings form a tailed distribution resembling a comet-like shape. Samples near the decision boundary share lexical and structural characteristics with fraudulent postings, whereas higher-confidence legitimate examples are pushed farther away, forming the tail. At the same time, the non-fraudulent cluster becomes increasingly compact, indicating that the pull mechanism is successfully reducing intra-class variation. A similar trend is observed for fraudulent postings, which evolve from a scattered distribution into a tighter and more coherent cluster as the push mechanism separates them from the legitimate manifold.

\subsubsection{\MakeUppercase{Topological Structure (t-SNE and UMAP)}}

We further analyze the learned embeddings using t-SNE and UMAP projections (Figure~\ref{fig:tsne_umap}). The t-SNE visualization reveals elongated and continuous structures within the fraudulent class, suggesting that fraudulent postings may vary along a spectrum of scam templates rather than forming a single homogeneous cluster. Dense regions correspond to frequently occurring patterns, while the continuous trajectories indicate gradual transitions between related fraud types.

UMAP produces a similar overall structure while providing clearer global separation between classes \cite{mcinnes2018umap}. Both legitimate and fraudulent postings exhibit internal sub-structures, reflecting fine-grained semantic variations within each class while maintaining a clear distinction between classes. These observations complement the quantitative clustering results and provide visual evidence that CGCL promotes both intra-class compactness and inter-class separability in the learned latent space.
\begin{table*}[!t]
\centering
\caption{Per-seed test-set results (F1 / PR-AUC / MCC). For each seed, best F1, best PR-AUC, and best MCC across variants are individually \underline{underlined}.}
\label{tab:ablation_perseed}
\setlength{\tabcolsep}{3pt}
\renewcommand{\arraystretch}{1.3}
\small
\begin{tabular}{l *{5}{C{0.75cm}C{0.75cm}C{0.75cm}}}
\toprule
& \multicolumn{3}{c}{\textbf{Seed 7}}
& \multicolumn{3}{c}{\textbf{Seed 13}}
& \multicolumn{3}{c}{\textbf{Seed 21}}
& \multicolumn{3}{c}{\textbf{Seed 42}}
& \multicolumn{3}{c}{\textbf{Seed 99}} \\
\cmidrule(lr){2-4}\cmidrule(lr){5-7}\cmidrule(lr){8-10}
\cmidrule(lr){11-13}\cmidrule(lr){14-16}
\textbf{Variant}
  & F1 & {\scriptsize PR-AUC} & MCC
  & F1 & {\scriptsize PR-AUC} & MCC
  & F1 & {\scriptsize PR-AUC} & MCC
  & F1 & {\scriptsize PR-AUC} & MCC
  & F1 & {\scriptsize PR-AUC} & MCC \\
\midrule
CE\_only
  & .727 & .741 & .716
  & .742 & .749 & .732
  & .689 & .702 & .675
  & .734 & .744 & .724
  & .696 & .699 & .681 \\
CE\_weighted
  & .751 & .701 & .741
  & .782 & .780 & .773
  & .732 & .719 & .718
  & .753 & .770 & .741
  & .702 & .733 & .688 \\
CE\_ADASYN
  & .980 & .982 & .972
  & \underline{.981} & .982 & \underline{.973}
  & .980 & .976 & .972
  & .981 & \underline{.986} & .974
  & .982 & .982 & .974 \\
CE\_CenterLoss
  & .978 & .975 & .970
  & .980 & .974 & .972
  & \underline{.984} & .981 & \underline{.977}
  & \underline{.983} & .981 & \underline{.977}
  & \underline{.984} & \underline{.991} & \underline{.978} \\
CGCL\_full
  & \underline{.984} & \underline{.990} & \underline{.978}
  & .980 & \underline{.988} & .972
  & .978 & \underline{.984} & .970
  & .981 & .984 & .974
  & .931 & .961 & .905 \\
\bottomrule
\end{tabular}
\end{table*}

% ===== Ablation study inserted from ablation Study.tex =====
\subsection{\MakeUppercase{Ablation Study}}
\subsubsection{Introduction}
Fraud detection suffers from severe class imbalance, which biases standard cross-entropy classifiers toward the majority (legitimate) class and suppresses fraud recall. This report evaluates five ablation variants on an identical MLP backbone, isolating the contribution of loss re-weighting, synthetic oversampling (ADASYN), center loss, and the full CGCL model. All variants share the same architecture, features, and optimizer; only the loss objective and sampling strategy differ. Results are averaged over five random seeds $\{7, 13, 21, 42, 99\}$.

\subsubsection{\MakeUppercase{Experimental Setup}}

\paragraph{Model and Training}
Each variant uses a three-hidden-layer MLP (256--128--64, ReLU + BatchNorm) trained with Adam ($\text{lr}=10^{-3}$, weight decay $10^{-4}$) for up to 500 epochs.

\paragraph{Ablation Variants}
\begin{itemize}\setlength\itemsep{2pt}
  \item \textbf{CE\_only} — Standard cross-entropy; no imbalance handling.
  \item \textbf{CE\_weighted} — Cross-entropy with inverse-frequency class weights.
  \item \textbf{CE\_ADASYN} — Cross-entropy on ADASYN-oversampled training data.
  \item \textbf{CE\_CenterLoss} — Cross-entropy + center loss ($\lambda=0.5$) to compact intra-class embeddings.
  \item \textbf{CGCL\_full} — Full proposed model: contrastive graph objective + cross-entropy.
\end{itemize}

\subsubsection{\MakeUppercase{Results}}

\paragraph{Quantitative Summary}

Table~\ref{tab:ablation_results} reports mean $\pm$ std of all metrics across five seeds. CE\_only and CE\_weighted achieve $\sim$97\% accuracy yet score below F1~=~0.75 on the fraud class---a result of majority-class bias. Every variant explicitly addressing imbalance exceeds F1~=~0.97.

\begin{figure*}[!t]
  \centering
  \SafeIncludeGraphics[width=0.84\textwidth]{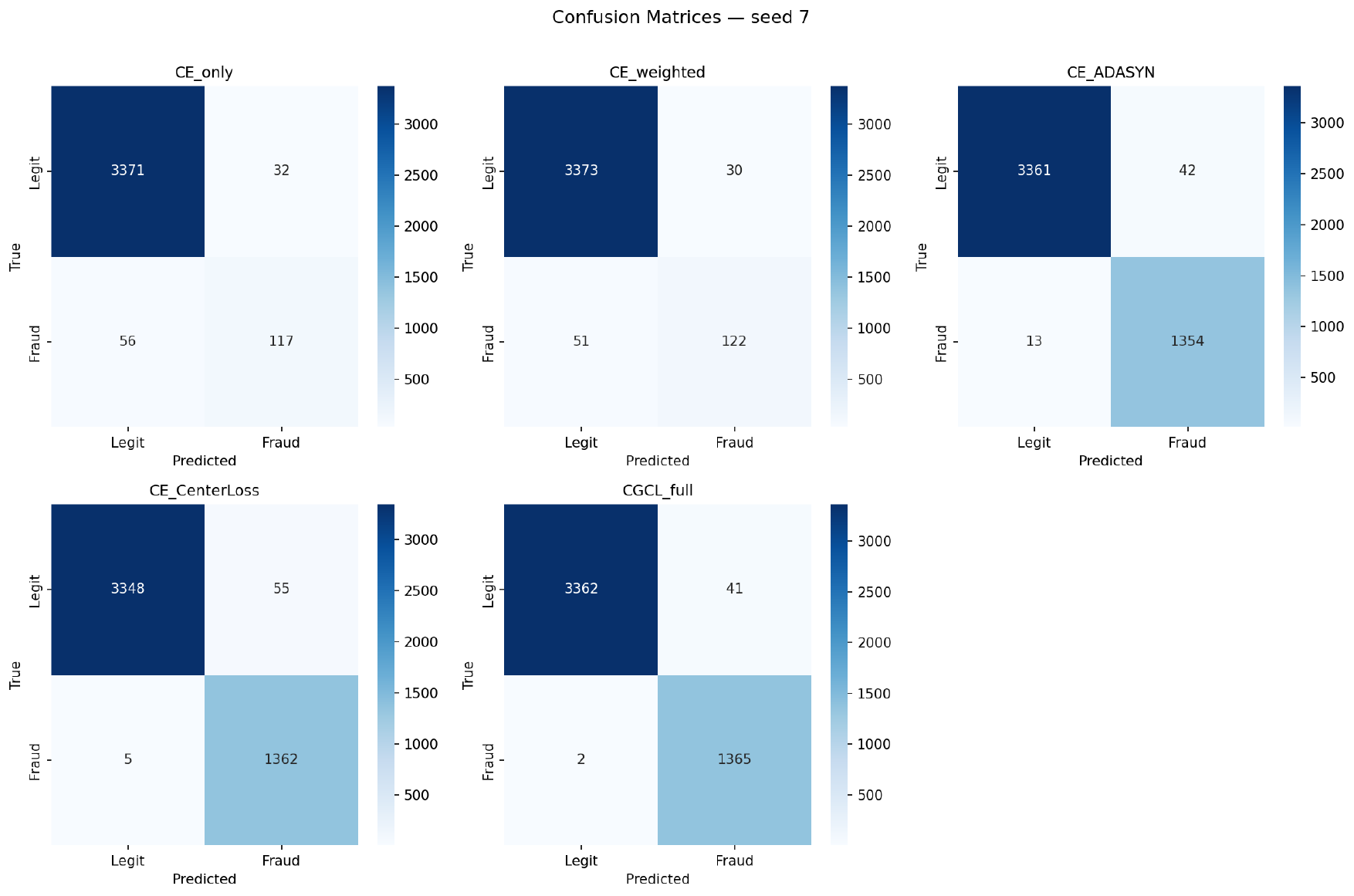}
  \caption{Confusion matrices for all five variants (seed~7). Top row (L$\to$R): CE\_only, CE\_weighted, CE\_ADASYN. Bottom row: CE\_CenterLoss, CGCL\_full.}
  \label{fig:ablation_confusion_matrices}
\end{figure*}

\paragraph{Confusion Matrices (Seed 7)}

Figure~\ref{fig:ablation_confusion_matrices} shows confusion matrices for seed~7. CE\_only and CE\_weighted produce 56 and 51 false negatives respectively. Enhanced variants improve sharply: CE\_ADASYN yields 13 false negatives, CE\_CenterLoss 5, and CGCL\_full only 2.

\subsubsection{\MakeUppercase{Discussion}}

\paragraph{Class imbalance as the primary bottleneck.}
The 26-point F1 gap between CE\_only (0.718) and CE\_ADASYN (0.981) with no architectural change confirms that the MLP backbone is not the bottleneck. Loss re-weighting alone yields only marginal improvement (+0.026 F1), consistent with prior findings that synthetic oversampling is more effective than static weights for highly skewed distributions.

\paragraph{Center loss stability.}
CE\_CenterLoss matches CE\_ADASYN in F1 and MCC while achieving the lowest seed variance across all metrics. The center-loss regularizer tightens the fraud-class embedding cluster, reducing false negatives (5 vs.\ 13 for ADASYN on seed~7) and stabilising the decision boundary across initializations.

\paragraph{CGCL graph context and variance.}
CGCL\_full achieves the joint-best Recall (0.997) and highest PR-AUC (0.981), and virtually eliminates false negatives (2 on seed~7). However, its F1 variance (±0.020) and MCC variance (±0.028) are notably higher than CE-based variants. Seed~99 converged significantly more slowly, suggesting sensitivity to the interaction between graph topology and the contrastive temperature hyper-parameter---an open problem for future work.

% ===== End inserted ablation study =====

\section{\MakeUppercase{Conclusion}}

In this paper, we introduced a Centroid-Guided Contrastive
Loss (CGCL) loss function that is capable of enforcing accurate decision boundaries while consistently restructuring the
latent space into dense and well-separated clusters. Our loss
function dynamically updates unique clusters during training
using centroid-guided push and mechanism where top-$k$ hard-positives are pulled toward the active cluster while top-$k$ hard-negatives are pushed farther away. We also integrate Cross-Entropy loss to complement the contrastive objective, ensuring
strong class discrimination alongside compact cluster formation. Extensive experiments demonstrate that our approach
achieves state-of-the-art (SOTA) performance in both classification and clustering metrics, highlighting the effectiveness of
CGCL as a unified loss function.

\bibliographystyle{IEEEtran}
\bibliography{r1}
\end{document}